\documentclass[runningheads]{llncs}

\usepackage{accv}

\usepackage{accvabbrv}
\usepackage{graphicx}
\usepackage{amsmath,amssymb}
\usepackage[table,xcdraw]{xcolor}
\usepackage{booktabs}
\usepackage{makecell}
\usepackage{colortbl}
\usepackage[absolute,overlay]{textpos}
\usepackage{caption}
\usepackage{subcaption} 
\usepackage{multirow}
\usepackage{array}
\usepackage{adjustbox}
\usepackage{tabularx}
\usepackage{float}
\usepackage{arydshln}
\usepackage{pifont}
\usepackage{wrapfig}
\usepackage{xspace} 

\usepackage[accsupp]{axessibility}  

\arrayrulecolor{black}

\newcommand{\hk}[1]{\textcolor{black}{#1}}

\usepackage[breaklinks,colorlinks,citecolor=blue,linkcolor=blue,urlcolor=blue]{hyperref}

\usepackage{orcidlink}

\begin{document}

\title{Industrial Synthetic Segment Pre-training}

\titlerunning{Industrial Synthetic Segment Pre-training}

\author{
Shinichi Mae\inst{1}\textsuperscript{\textdagger} \and
Hirokatsu Kataoka\inst{2,3}\textsuperscript{\textdagger} \and
Ryousuke Yamada\inst{2,4} \and \\
Yoshihiro Fukuhara\inst{2,5} \and
Risa Shinoda\inst{6} \and
Christian Rupprecht\inst{3}
}

\authorrunning{S.~Mae et al.}

\institute{
Toyota Industries Corporation \and
National Institute of Advanced Industrial Science and Technology (AIST) \and
Visual Geometry Group, University of Oxford \and
University of Technology Nuremberg \and
CADDi \and
The University of Tokyo
}
\maketitle
\begingroup
\renewcommand{\thefootnote}{\textdagger}
\footnotetext{These authors contributed equally.}
\endgroup
\begin{abstract}
Vision Foundation Models (VFMs) have made remarkable progress and are increasingly being applied to segmentation tasks in real-world industrial settings. However, VFMs pre-trained on real-image datasets still face several challenges: (1) they do not always perform well on industrial datasets due to significant differences from natural imagery, (2) legal and ethical restrictions, such as limitations on commercial use, constrain extensibility, and (3) building training frameworks under limited computational and data resources remains a critical issue. These challenges raise a fundamental question: can we construct industrial segmentation models without relying on real images or manual annotations? To address this question, we propose the Instance Core Segment Dataset (InsCore), a synthetic data generation framework and the resulting pre-training dataset based on Formula-Driven Supervised Learning (FDSL). InsCore is designed not around the visual appearance or domain of real images, but around the hypothesis that learning to handle complex occlusions during pre-training is a key factor for strong performance in industrial domains. Through experiments across five domains (medical, biomedical, remote sensing, manufacturing, and logistics) we demonstrate that InsCore pre-trained models achieve average mAP scores of 45.2 with the ViTDet backbone and 46.0 with the Swin Transformer backbone, on par with ImageNet-21k supervised pre-training (45.0) while using no real images at all. As a reference point under different input assumptions, prompted SAM with ground-truth bounding boxes attains 45.4 on the same benchmarks. Finally, InsCore consists of only 100k images and 3.2M masks, roughly 1/110 and 1/312 the scale of the SA-1B dataset.
\end{abstract}

\section{Introduction}
\label{sec:intro}
In recent years, the field of computer vision has entered the era of foundation models, characterized by very large-scale pre-trained models~\cite{dosovitskiy2021vit, zhai2022scalingvit, radford2021clip}. Both fully- and self-supervised ViT backbone models, such as the Segment Anything Model (SAM)~\cite{Kirillov_2023_ICCV,ravi2024sam2} and the DINO series~\cite{caron2021emerging, oquab2024dinov2, simeoni2025dinov3}, have quickly emerged, demonstrating broad abilities in segmentation tasks. However, most of these vision foundation models (VFMs) are trained on extremely large image datasets

\begin{wrapfigure}[32]{r}{0.5\textwidth}
    \centering
    \begin{subfigure}{\linewidth}
        \centering
        \vspace{-10pt}
        \includegraphics[width=\linewidth]{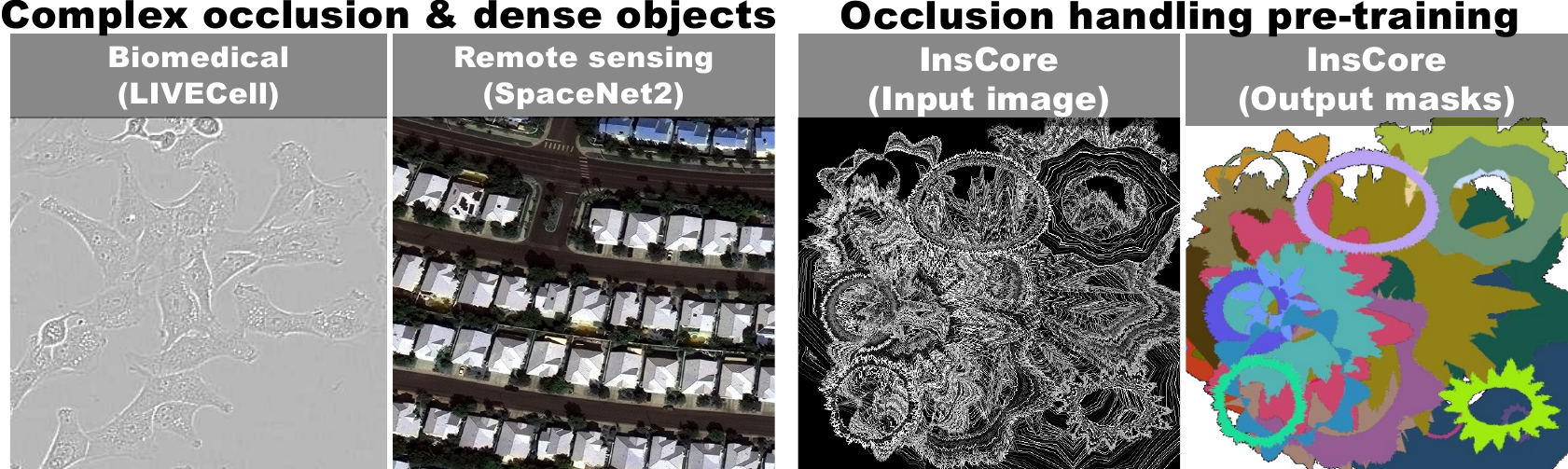}
        \caption{Our instance core segment dataset (InsCore) inspired by industrial data. While visual appearance and domain gaps between web and industrial data are frequently discussed, our working hypothesis is that a critical gap lies in \textbf{occlusion handling} during instance segmentation. We investigate whether a pre-training task designed around this hypothesis can match real-image pre-training on industrial datasets.}
        \label{fig:inscore_industrial_data}
        
    \end{subfigure}
    \begin{subfigure}{\linewidth}
        \centering
        \includegraphics[width=\linewidth]{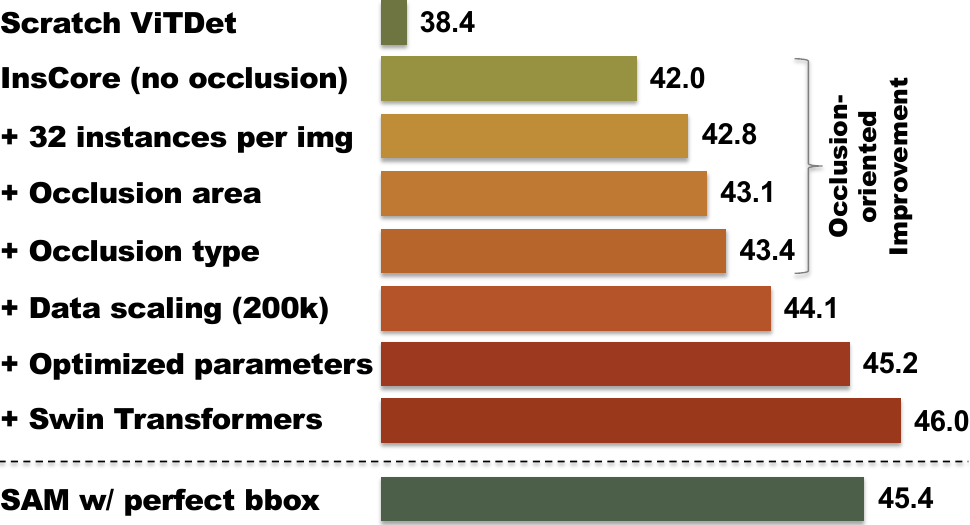}
        \caption{We propose the \textbf{Instance Core Segment Dataset (InsCore)}, a commercially usable synthetic pre-training dataset for industrial instance segmentation. InsCore synthesizes complex occlusions and dense, hierarchical masks---key characteristics of industrial data---using a single formula, addressing the commercial-use restrictions and limited industrial transferability of datasets such as ImageNet and SA-1B. InsCore delivers pre-training performance on par with ImageNet-21k, and is competitive with prompted SAM, reported as a reference point under different input assumptions.}
        \label{fig:inscore_upgrade}
    \end{subfigure}
    \caption{Our motivation, pre-training task, and results for InsCore-based synthetic pre-training.}
    \label{fig:figone}
\end{wrapfigure}

\noindent
 and their use often comes with licensing restrictions or opaque terms that complicate commercial adoption\footnote{SAM has been released under the Apache~2.0 license, allowing for commercial use. However, the SA-1B was provided under the research purposes only, and the company can gradually add usage restrictions.}. Moreover, the enormous computational resources required for the VFMs mean there remains a significant gap between academic research and industrial application~\cite{ji2024segment, Zhang_2024_CVPR}. For instance segmentation, the task of delineating each object at the pixel level, widely used in industrial applications such as medical imaging and remote sensing, constructing large-scale real-world industrial datasets is challenging. Directly applying web-trained general models to specialized industrial data remains problematic.

Even recent VFMs such as SAM have shown only limited gains on industrial datasets~\cite{ji2024segment}, mainly because they were trained on generic web images, whose accuracy degrades on domain-specific data such as medical or satellite imagery. In industrial settings where data quantity and usage rights are tightly constrained, there is a need for approaches that achieve high performance without relying on proprietary large-scale data or restrictive licenses.

To address these challenges, we turn to synthetic data that can be used commercially without restrictions. Conventional methods using CG or CAD tools can automatically generate images with segmentation labels~\cite{tobin2017domain}, avoiding manual annotation; however, such synthetic images often fail to capture the complexity of real-world data, yielding only limited performance gains. Simply mimicking visual appearance is insufficient; a more effective strategy is to derive fundamental feature representations from first principles. One such approach is \hk{Formula-Driven Supervised Learning} (FDSL)~\cite{kataoka2020pre}, which captures the principled features of recognition models like Vision Transformer (ViT)~\cite{dosovitskiy2021vit} during pre-training. Originally applied to image classification, FDSL using contour-based shape projections outperformed ImageNet-21k~\cite{deng2009imagenet} and achieved accuracy closely comparable to JFT-300M~\cite{sun2017revisiting} using only synthetic data~\cite{takashima2023visual}. Inspired by industrial data (Figure~\ref{fig:inscore_industrial_data}), we argue that if a pre-training task is designed to effectively capture complex occlusions in instance segmentation, synthetic pre-training can outperform both conventional synthetic methods and real-image pre-training.

In this work, we propose the \hk{Instance Core Segment Dataset (InsCore)}, a synthetic data generation framework and the pre-training dataset it produces for instance segmentation, which generates \hk{complex occlusion patterns with precise annotations} (Figure~\ref{fig:inscore_industrial_data}); throughout the paper, we use the name InsCore for both the framework and the resulting dataset, with the intended meaning clear from context. By overlaying many clutter patterns in an image, InsCore simulates complex, overlapping scenarios and produces precise pixel-level annotations automatically. This yields a license-free dataset without any data collection, manual labeling, or cross-checking at any phase. Experiments show that models pre-trained on InsCore achieve average performance on par with widely used real-image pre-training (ImageNet-21k) across five representative industrial datasets spanning medical, biomedical, remote sensing, manufacturing, and logistics domains, and clearly outperform existing synthetic pre-training datasets (Figure~\ref{fig:inscore_upgrade}). We additionally report prompted SAM and SAM2 as reference points under different input assumptions, and find that InsCore pre-training is competitive with them on average. Notably, InsCore pre-training uses only 100k-200k synthetic images, 50-100$\times$ fewer than the 11 million images used to train SAM.

\noindent\textbf{Paper contribution.}
(i) We present InsCore, an industrial synthetic segment pre-training framework that automatically generates synthetic images with complex occlusions along with precise pixel-level annotations. (ii) A ViT pre-trained on InsCore matches ImageNet-21k supervised pre-training on five industrial datasets without any real images, and achieves average accuracy comparable to prompted SAM provided with ground-truth bounding boxes, despite solving the harder task of jointly estimating both location and category from the image alone. (iii) InsCore requires no manual effort, circumventing the legal and ethical restrictions of real-image datasets and providing a pre-training dataset that is fully ready for commercial use. Moreover, InsCore is compact\hk{: it contains 100k images and 3.2M masks, roughly 110$\times$ and 312$\times$ fewer than SA-1B (11M images, 1B masks), although the two datasets target different purposes (Section~\ref{sec:toward_industrial_vfm})}.

\section{Related work}
\label{sec:related work}
\noindent{\textbf{Instance segmentation for industrial applications.}}  
Instance segmentation, essential for industrial applications, requires precise boundary delineation of each object within an image~\cite{he2017mask, liu2021swin, bolya2019yolact, cheng2022masked}.  
Recognition performance has improved significantly through pre-training on large-scale datasets such as ImageNet~\cite{deng2009imagenet} and COCO~\cite{lin2014microsoft}.  
However, models such as the SAM series~\cite{Kirillov_2023_ICCV,ravi2024sam2,carion2025sam3segmentconcepts}, which perform well on general-purpose datasets such as ADE20K~\cite{zhou2017scene}, often show suboptimal results in industrial contexts due to mismatches in data distribution~\cite{osco2023segment, he2023computer}.  
In addition, concerns related to privacy, copyright, and social bias in real-image datasets~\cite{yang2019fairer,birhane2023into} present further obstacles.  

A more fundamental issue is that standard datasets such as ImageNet and COCO are not permitted for commercial use. 
Nevertheless, legal and ethical challenges associated with real-image usage remain unresolved.  
This study investigates an alternative approach \hk{that avoids real images and manual labeling,} building high-precision models for industrial applications through synthetic pre-training, entirely independent of real images.

\noindent{\textbf{Synthetic image generation for image recognition.}}  
One promising approach to resolving commercial usage rights issues while achieving performance comparable to real-data methods is the use of synthetic data for pre-training, which has gained attention as a cost-effective alternative to manual dataset construction~\cite{chen2019learning,mishra2022task2sim,michieli2020adversarial,tremblay2018training,billot2023synthseg,tian2024stablerep}.  
For example, Task2Sim enhances image recognition by manipulating factors such as lighting and object size during pre-training~\cite{mishra2022task2sim}, and more recent work leverages photorealistic images generated by models like Stable Diffusion~\cite{tian2024stablerep} or synthesizes images from generated segmentation masks (e.g., SegGen).  
However, these methods can struggle to capture real-world complexity due to inconsistencies in generated data and high computational costs~\cite{ye2024seggen}.  

In contrast, formula-driven synthetic training has been proposed to effectively overcome copyright and ethical issues while maintaining high accuracy~\cite{kataoka2020pre,takashima2023visual,shinoda2023segrcdb,Kataoka_2021_ICCV,yamada2021mv,Kataoka_2022_CVPR,yamada2022pcfractaldb,tadokoro2023pre,yamada2024fsvgp,nakashima2022vitfractaldb,nakamura2023pre,nakamura2024scalingbackwards}. This framework automatically generates synthetic images based on natural laws and has proven effective in uncovering fundamental principles of deep learning.
A notable derivative, SegRCDB~\cite{shinoda2023segrcdb}, is designed for semantic segmentation and achieves performance comparable to some real and synthetic datasets.
We advocate constructing synthetic datasets grounded in essential principles to address critical issues of data licensing and domain transferability in industrial contexts\hk{, and further explore in-depth occlusion handling pre-training in instance segmentation}.  
This approach is key to bridging the gap between academic methods and real-world deployment.

\noindent{\textbf{VFMs for segmentation tasks.}}
In recent years, numerous VFMs with advanced visual capabilities have emerged, particularly those that achieve remarkable performance in segmentation tasks. 
Among them, representative examples include the segment anything models (SAM, SAM2, and SAM3)~\cite{Kirillov_2023_ICCV,ravi2024sam2,carion2025sam3segmentconcepts} and the DINO series (especially DINOv2 and DINOv3)~\cite{caron2021emerging, oquab2024dinov2, simeoni2025dinov3}, which are frequently referenced in recent studies. 
However, it has been noted that SAM does not necessarily achieve perfect segmentation for all scenes or objects in the real world~\cite{ji2024segment}. Considering practical applications, we believe that developing models with direct industrial segmentation capabilities for real-world objects is crucial, and this effort will further contribute to the functional expansion of VFMs.
We therefore investigate whether an industrial segmentation model can be constructed that achieves performance comparable to real-image pre-training and to prompted VFMs without real images, without human supervision, under realistic computational resources (8 GPUs in a single node), and using a standard model architecture (ViT backbone with ViTDet~\cite{li2022exploring}).

\begin{table}[t]
\centering
\caption{
Industrial datasets used to evaluate pre-training effects, along with their domains, training data and mask counts (\#Train images), test data and mask counts (\#Test images), and number of classes (\#Categories).
}
\resizebox{\textwidth}{!}{
\begin{tabular}{llrrrrc}\toprule[0.8pt]
\multirow{2}{*}{Evaluation dataset} & 
\multirow{2}{*}{Industrial domain} &
\multicolumn{2}{c}{\#Train \hk{set}} &
\multicolumn{2}{c}{\#Test images}  &
\multirow{2}{*}{\#Categories} \\
& & \hk{\#Images} & \hk{\#Masks} & \hk{\#Images} & \hk{\#Masks} &  \\
\midrule[0.5pt]
Endoscapes (ES)~\cite{murali2023endoscapes} & Medical & 343 & 1,615 & 74 & 270 & 6 \\
LIVECell (LC)~\cite{edlund2021livecell} & Biomedical & 3,253  & 1,018,576 & 1,564  & 462,261 & 1 \\
SpaceNet2 (SN)~\cite{van2018spacenet}  & Remote sensing  & 3,080  & 87,301   & 771   & 21,641  & 1 \\
Industrial-iSeg (IS)~\cite{Li_Wong_2024} & Manufacturing   & 1,109  & 25,308   & 89    & 523    & 6 \\
LogiSeg (LS)~\cite{mae2025efficient}     & Logistics       & 1,384  & 10,018   & 300   & 2,093   & 7 \\
\midrule[0.3pt]
\multicolumn{7}{c}{\includegraphics[width=1.2\textwidth]{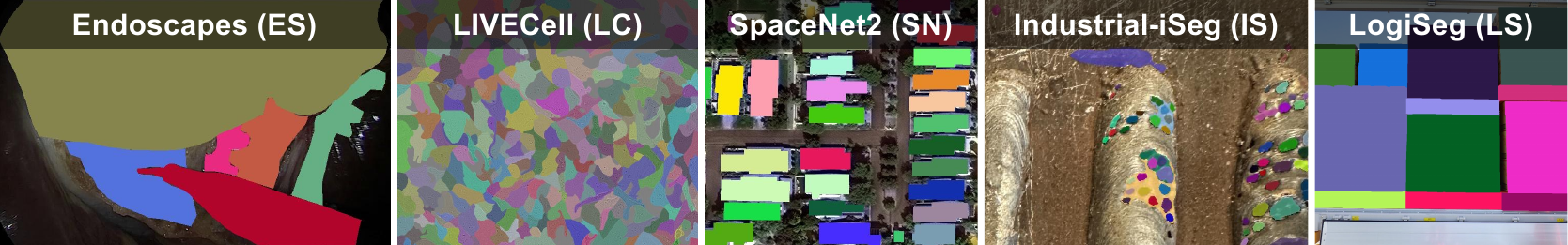}} \\
\bottomrule[0.8pt]
\end{tabular}
}
\label{tab:datasets}
\end{table}

\section{Instance core segment dataset (InsCore)}
\label{sec:method}

\subsection{Why is InsCore effective for industrial data?}
\label{sec:motivation}

\noindent\textbf{Complex occlusion is a key challenge on industrial datasets.}
Instance segmentation in industrial applications differs fundamentally from that in general-purpose web datasets, particularly in terms of complex occlusions and dense object arrangements.
While visual appearance and domain gaps are frequently discussed, our working hypothesis is that occlusion handling is an important factor for instance separation in such settings.
In many industrial scenes, objects may provide fewer distinctive texture cues (e.g., similar materials or low intra-class texture variation), making boundary delineation under severe occlusion a critical factor for separating instances.
Industrial datasets also demand high annotation precision and robustness to partial visibility, and heavily occluded scenes are painstaking to annotate manually.
We note that occlusion handling is intertwined with dense, pixel-accurate annotation and boundary learning; we analyze occlusion-related design choices in Section~\ref{sec:ablation_inscore} and discuss open questions in Section~\ref{sec:toward_industrial_vfm}.

Table~\ref{tab:datasets} describes and illustrates the industrial datasets for instance segmentation used in this paper.
Endoscapes~\cite{murali2023endoscapes} features complex occlusions caused by organs and surgical instruments, and LIVECell~\cite{edlund2021livecell} contains large numbers of overlapping cells per image; accurate segmentation in such scenes is often difficult even for domain experts due to ambiguous boundaries and frequent occlusions.
SpaceNet2~\cite{van2018spacenet} and LogiSeg~\cite{mae2025efficient} contain densely packed, high-precision masks in crowded scenes, while Industrial-iSeg~\cite{Li_Wong_2024} contains fewer occlusions but still demands precise annotation.
Together, these datasets motivate pre-training data that emphasizes accurate boundary learning under occlusion and dense layouts.

\noindent\textbf{Industrial synthetic segment pre-training.}
We construct a synthetic pre-training dataset based on two essential properties: (i) complex occlusions and (ii) perfectly accurate, pixel-level annotations.
To address this, we propose InsCore, a synthetic data generation framework and its resulting pre-training dataset, based on FDSL, a methodology for constructing pre-training datasets from generative rules.
InsCore generates hollow-shaped masks defined as the region between the inner and outer contours of a single formula-driven shape (Figure~\ref{fig:method}).
This hollow design encourages the network to learn feature representations of both inner and outer object boundaries, which can reflect interlocking and nested structures observed in industrial components (e.g., machine parts) and biological structures (e.g., overlapping cells).
By placing these shapes within an image, InsCore simulates complex occlusions with perfectly accurate pixel-level instance annotations. Each shape is treated as an individual instance mask, enabling fully supervised instance-level learning.

We additionally describe how to construct a basic dataset for instance segmentation in Section~\ref{sec:generation} and complex occlusion handling in Section~\ref{sec:occlusion_handling}.

\begin{wrapfigure}[16]{r}{0.5\textwidth}
  \centering
  \vspace{-10pt}
  \includegraphics[width=\linewidth]{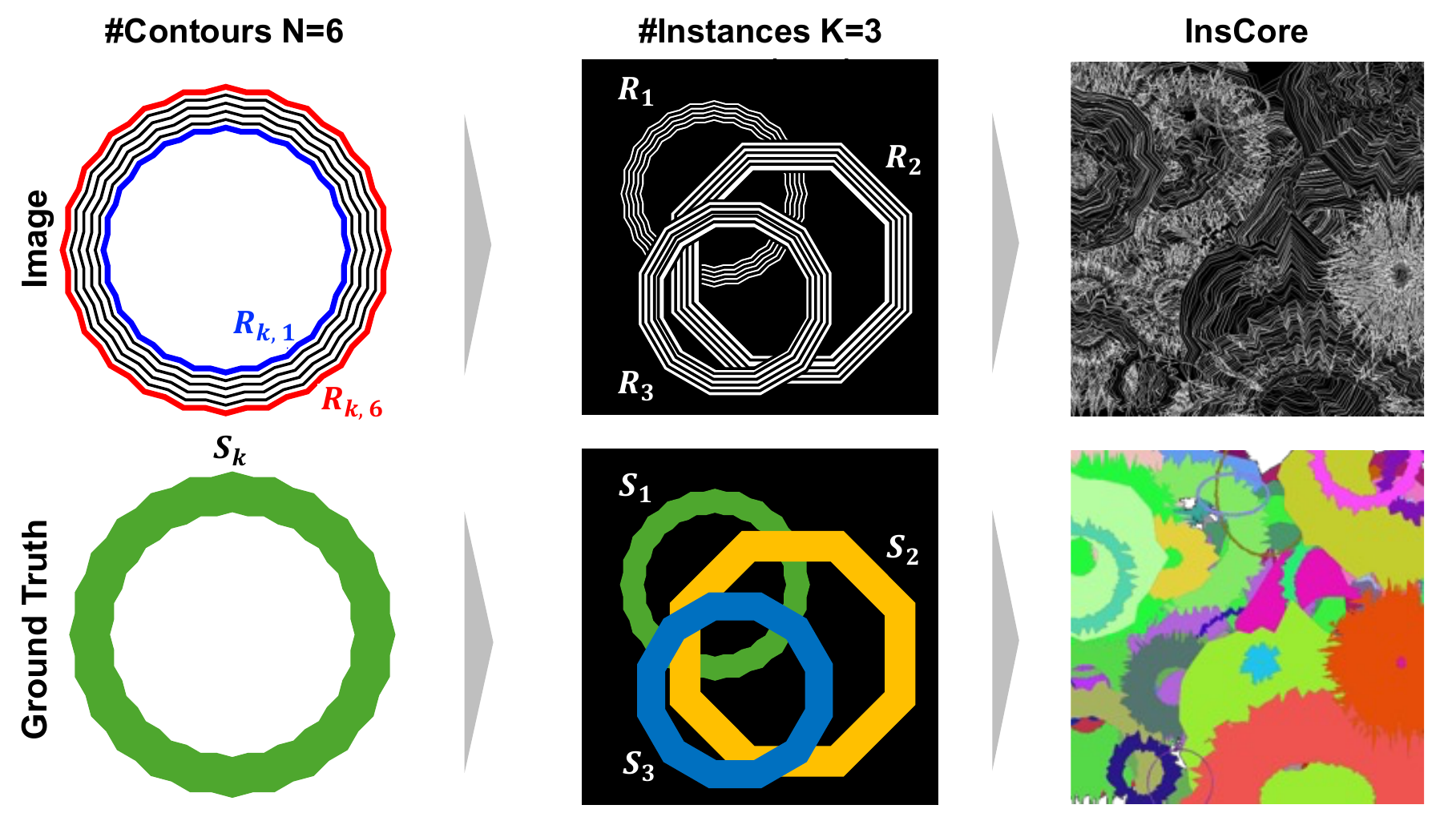}
  \caption{InsCore enables synthesizing complex occlusions and their masks observed in industrial data by placing hollow masks $S_k$, defined as the region between the innermost and outermost contours of formula-generated image patterns.}
  \label{fig:method}
\end{wrapfigure}

\subsection{Generation process in InsCore (Figure~\ref{fig:method})}
\label{sec:generation}

\noindent\textbf{Rendering contour shapes.}

\noindent In InsCore, the discrete image canvas is defined as a two-dimensional grid
$\Omega_d = \{0, \dots, W-1\} \times \{0, \dots, H-1\}$,
where $W$ and $H$ denote the width and height of the image, respectively.
The image is defined as a mapping $x : \Omega_d \rightarrow [0, 255]^3$ to allow for continuous rendering values such as anti-aliasing, and is initialized with a black background,
$x^{(0)}(u, v) = \mathbf{0}$.
The number of instances per image, $K_i$, varies for each sample and is determined as a positive integer sampled from a discrete uniform distribution
$\mathcal{U}_{\mathbb{Z}}(1, K_{\text{max}})$, where $K_{\text{max}} = 64$.

Each instance $O_k$ ($k = 1, \dots, K_i$) is represented by a contour shape $\mathcal{R}_k \subset \mathbb{R}^2$ generated based on the RCDB formulation.
We introduce a continuous spatial domain $\Omega = [0, W) \times [0, H)$ for describing vertex coordinates.
The instance is generated using a parameter set
$\eta_k = (N_k, n_k, r_k, \Delta_k, lw_k, \mathbf{o}_k, \boldsymbol{\lambda}_k, \mathbf{p}_k)$,
where $\mathbf{p}_k \in \mathbb{R}^2$ is the instance center.
The contour shape is defined as a union of polygonal edges:
\begin{equation}
\mathcal{R}_k = \bigcup_{p=1}^{N_k} R_{k,p},
\quad
R_{k,p} = \bigcup_{j=0}^{n_k-1} e\!\left(\mathbf{v}_{j}^{(k,p)}, \mathbf{v}_{(j+1) \bmod n_k}^{(k,p)}\right).
\end{equation}
Each edge $e(\cdot,\cdot)$ is defined as a linear interpolation between vertices, offset by the instance center position $\mathbf{p}_k$, as follows:
\begin{equation}
e(\mathbf{a}, \mathbf{b}) =
\left\{
(1 - t)\mathbf{a} + t\mathbf{b} + \mathbf{p}_k \in \mathbb{R}^2
\ \middle|\ 
0 \le t \le 1
\right\}.
\end{equation}
For the initial shape, the first polygon $R_{k,1}$ is constructed as a regular $n_k$-gon with radius $r_k$ and aspect ratio $\mathbf{o}_k = (o_{x,k}, o_{y,k})$:
\begin{equation}
\mathbf{v}_{j}^{(k,1)} = r_k
\begin{pmatrix}
o_{x,k} \cos\left( \frac{2\pi j}{n_k} \right) \\
o_{y,k} \sin\left( \frac{2\pi j}{n_k} \right)
\end{pmatrix},
\quad
j = 0, \dots, n_k-1.
\end{equation}
Subsequent polygons $R_{k,p}$ ($p \geq 2$) are recursively generated by radially expanding the vertices of $R_{k,p-1}$, perturbed by Perlin noise.
To reduce the risk of topological nesting failures caused by extreme negative expansions, the radial increments are strictly clipped to non-negative values using a base expansion step $\Delta_k$:
\begin{equation}
\mathbf{v}_{j}^{(k,p)} = \mathbf{v}_{j}^{(k,p-1)} +
\begin{pmatrix}
\max\!\bigl(0, \Delta_k + \lambda_{x,k}\, \epsilon_{j,p-1}\bigr)\, \cos\left( \frac{2\pi j}{n_k} \right) \\
\max\!\bigl(0, \Delta_k + \lambda_{y,k}\, \epsilon_{j,p-1}\bigr)\, \sin\left( \frac{2\pi j}{n_k} \right)
\end{pmatrix}.
\end{equation}
Here, $\epsilon_{j,p-1} \in [-1, 1]$ is a sample from a one-dimensional Perlin noise sequence, and $\boldsymbol{\lambda}_k = (\lambda_{x,k}, \lambda_{y,k})$ are noise scaling factors.
All instances are rendered sequentially from back ($k=1$) to front ($k=K_i$).
The intermediate image $x^{(k)}$ is updated by a rendering operator $\mathrm{Render}$ that draws the contour $\mathcal{R}_k$ as white lines with a stroke width $lw_k$, which is expressed as $x^{(k)} = \mathrm{Render}(x^{(k-1)}, \mathcal{R}_k, lw_k)$.
Let $\mathcal{A}_d(R)$ denote the rasterized 2D filled region bounded by a closed contour $R$ on the discrete grid $\Omega_d$, defined using the standard even-odd filling rule.
For each instance, we construct the outermost filled region $\mathcal{A}_d(R_{k,N_k})$ and the innermost filled region $\mathcal{A}_d(R_{k,1})$.
To guarantee topological validity, we strictly enforce
$\mathcal{A}_d(R_{k,1}) \subset \mathcal{A}_d(R_{k,N_k})$.
If this condition is violated, the instance parameters are resampled.
These regions are then used for ground-truth mask construction and occlusion processing.

\noindent\textbf{Instance segmentation masks with complex occlusions.}
Each contour instance $O_k$ consists of a nested sequence of polygons $\{R_{k,p}\}_{p=1}^{N_k}$.
The segmentation mask region $S_k$ is defined as the hollow area between the outermost and innermost boundaries, given by $S_k = \mathcal{A}_d(R_{k,N_k}) \setminus \mathcal{A}_d(R_{k,1})$.
This hollow structure allows background instances to remain visible through the inner regions of foreground objects (i.e., the inner hole itself is transparent).

To simulate realistic occlusions, the final ground-truth instance mask includes only the visible regions of each instance.
Instances are rendered from back to front.
We define occlusion as being caused by the \emph{opaque} mask regions $S_i$ of subsequently rendered (foreground) instances (the hollow holes do not occlude).
The visible region $V_k$ of instance $O_k$ is obtained by subtracting these foreground mask regions, formulated as $V_k = S_k \setminus \bigcup_{i=k+1}^{K_i} S_i$.

Fully occluded pixels are excluded from the labels.
For instance segmentation, the binary mask $y_k(u,v) \in \{0, 1\}$ for each instance $k$ is formally defined as $y_k(u,v) = \mathbf{1}[(u,v) \in V_k]$.

Additionally, a semantic class map $m(u,v)$ can be constructed by assigning the uniformly sampled category label $c_k \in \{1, \dots, C\}$ to the mutually disjoint visible regions $V_k$:
\begin{equation}
m(u, v) =
\begin{cases}
c_k & \text{if } (u,v) \in V_k \text{ for some } k, \\
0 & \text{otherwise.}
\end{cases}
\end{equation}
Repeating this process for all $k = 1, \dots, K_i$, InsCore generates ground-truth instance masks that accurately model complex occlusion relationships in synthetic data.

\noindent\textbf{Data generation hyperparameters and sampling strategy.}
Each instance consists of $N_k$ polygons, where $N_k$ is uniformly sampled from $\{2, 3, \dots, 50\}$ (so that the hollow mask is non-empty), and each polygon contains $n_k$ vertices with $n_k \in \{3, \dots, 502\}$, enabling high shape diversity.

The center of each instance, $\mathbf{p}_k = (p_{x,k}, p_{y,k})$, is sampled to control overlap and boundary truncation.
By default, we sample $\mathbf{p}_k \sim \mathcal{U}(0, W) \times \mathcal{U}(0, H)$ over the image domain $\Omega$, naturally introducing overlap and occlusion.

Instance category labels $c_k$ are sampled uniformly from a class set $\mathcal{C} = \{1, \dots, 256\}$ (i.e., $C=256$), ensuring semantic diversity.
The remaining rendering parameters, including the stroke width $lw_k$, are randomly sampled per instance; their exact sampling ranges are specified in the released generation code to ensure full reproducibility.

All images are rendered as square RGB images with a resolution of $W = H = 512$, and the corresponding instance masks are generated at the same resolution.

\subsection{Occlusion handling in InsCore}
\label{sec:occlusion_handling}

\noindent\textbf{Number of instances per image (Table~\ref{tab:ablation_instances}; parameter $K$).}
Higher instance density increases the probability of inter-instance overlap, naturally generating complex occlusion patterns where multiple objects mutually occlude one another.
We validate $K$ from 1 to 64 instances per image.

\noindent\textbf{Occlusion density control (Table~\ref{tab:ablation_occlusion_dense}; parameters $K$ and $L$).}
To systematically control occlusion complexity and boundary truncation, we control the placement area size $L \in \{256, 512, 1024\}$ [pixel] used to sample instance centers.
Specifically, we define a placement domain centered at the image center,
\begin{equation}
\Omega_{\text{place}}(L)
=
\left[\frac{W}{2}-\frac{L}{2}, \frac{W}{2}+\frac{L}{2}\right)
\times
\left[\frac{H}{2}-\frac{L}{2}, \frac{H}{2}+\frac{L}{2}\right),
\end{equation}
and sample $\mathbf{p}_k \sim \mathcal{U}\!\bigl(\Omega_{\text{place}}(L)\bigr)$.
When $L>W(=H)$, instance centers can lie outside the image canvas, deliberately inducing boundary truncation where instances are partially cropped by image edges.
Combined with up to 64 overlapping instances per image, this configuration generates both inter-instance occlusion and boundary truncation, creating a challenging pre-training dataset that encourages robust feature learning for partially visible objects in cluttered scenes.

\noindent\textbf{Occlusion types (Table~\ref{tab:ablation_type}; parameter: placement strategy).}
To investigate the impact of spatial distributions on model generalization, we implement six instance placement strategies and evaluate their effects. See Figure~\ref{fig:occlusion_types} in the Appendix for visualizations.

\noindent\textit{(i) Full random.}
Each instance center is independently sampled from a uniform distribution over $\Omega$, serving as the baseline placement strategy.

\noindent\textit{(ii) Grid.}
Instances are arranged on a regular grid of size $\lceil\sqrt{N}\rceil$ (with $N$ instances), with random perturbations of $\pm$20\% of the cell size added to each position; positions are shuffled to prevent sequential ordering artifacts.

\noindent\textit{(iii) Gaussian.}
Instances follow a Gaussian distribution centered on the image with a standard deviation of 25\% of the image size.

\noindent\textit{(iv) Poisson.}
Instances are positioned via Poisson disk sampling with a minimum inter-instance distance of $\text{size}/10$, encouraging spatial uniformity.

\noindent\textit{(v) Cluster.}
Instances are organized into 2--5 clusters ($N/8$), whose centers are randomly placed with a 15\% margin from image boundaries; instances within each cluster follow a Gaussian with $\sigma \in [\text{size}/12, \text{size}/6]$.

\noindent\textit{(vi) Spiral.}
Instances follow a Fibonacci (golden-angle) spiral, with the $i$-th instance at angle $\theta_i = i \cdot \pi(3-\sqrt{5})$ and radius $r_i = r_{\text{max}}\sqrt{i/N}$, plus small random perturbations.

\noindent\textit{Mixed all.}
A balanced combination dataset created by uniformly sampling from all six placement strategies.
In our evaluation (Table~\ref{tab:ablation_type}), \textit{Full random} achieves the highest average score, and we adopt it as the default placement strategy for InsCore throughout the main experiments.
\textit{Mixed all} attains comparable performance and can serve as a robust alternative when the spatial priors of a target domain are unknown, since sampling from all strategies prevents the pre-trained model from overfitting to any single spatial prior.

\section{Experiments}
We compare our synthetic pre-training approach (InsCore) against established baselines using supervised and synthetic pre-training, including ImageNet-21k~\cite{deng2009imagenet} and existing synthetic datasets~\cite{roberts2021hypersim,cabon2020virtual,hu2019sail,shinoda2023segrcdb}. 
Additionally, to contextualize our results, we include VFMs trained on large-scale real images, such as SAM~\cite{Kirillov_2023_ICCV} and SAM2~\cite{ravi2024sam2}, as reference points. 
We validate the pre-training effectiveness on different types of industrial datasets used in the context of instance segmentation.

\subsection{Experimental settings}
\label{sec:experimental_settings}
\noindent\textbf{Industrial datasets.}
As described in Section~\ref{sec:motivation} \hk{and Table~\ref{tab:datasets}}, we evaluate the pre-training effectiveness on five industry-specific datasets covering diverse visual and structural challenges with complex occlusions and precise segmentation annotations: the medical (Endoscapes; laparoscopic cholecystectomy frames with tool--tissue occlusions~\cite{murali2023endoscapes}), biomedical (LIVECell; phase-contrast microscopy of densely packed cells~\cite{edlund2021livecell}), remote sensing (SpaceNet2; building footprints in satellite imagery, using the Las Vegas subset~\cite{van2018spacenet}), manufacturing (Industrial-iSeg; production-line defects such as welding faults and scratches~\cite{Li_Wong_2024}), and logistics (LogiSeg; loads and pallets in warehouses and truck yards~\cite{mae2025efficient}) domains.

\noindent\textbf{Implementation details.}
To evaluate instance segmentation performance with InsCore pre-training, we primarily employ ViTDet~\cite{li2022exploring}, specifically with a ViT-Base (ViT-B) backbone. \hk{ViTDet is primarily employed as a detection model; however, it uses Mask R-CNN~\cite{he2017mask} as a head module, allowing us to easily train an instance segmentation model with additional mask labels.}
We trained with the AdamW optimizer with an initial learning rate of 1e-4 and a weight decay of 0.1, applying a layer-wise learning rate decay of 0.7 to the ViT-B backbone. The model was trained for 100 or 200 epochs with a global batch size of 8 or 16, using mixed precision training for efficiency. All experiments used 8 H200 GPUs on a single node. 
We further investigate improved parameters for automatic synthetic data generation and InsCore pre-training in Section~\ref{sec:ablation_inscore}.

For SAM and SAM2, as described in the detailed comparison in the Appendix (Tables~\ref{tab:comparison_sam_finetuning} and~\ref{tab:sam_head_bboxnoise_investigation}), using ground-truth bounding boxes as prompts yields the best baseline performance.
\begin{wrapfigure}[9]{r}{0pt}
  \centering
  \includegraphics[width=0.42\columnwidth]{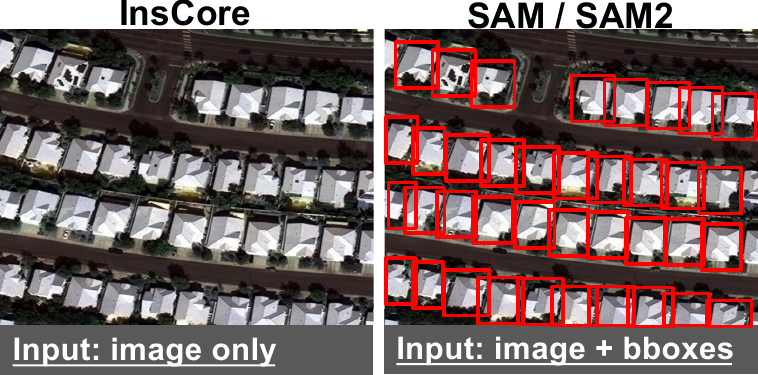}
  \caption{Input.}
  \label{fig:vit_sam_inputs}
\end{wrapfigure}
Ground-truth bounding boxes with small random perturbations \hk{(1-3 pixels; see Table~\ref{tab:sam_head_bboxnoise_investigation})} were provided as prompts to assess robustness under realistic conditions. 
\hk{Because SAM and SAM2 are class-agnostic, each mask predicted from a ground-truth box prompt inherits the category label of that box with confidence 1.0, and mAP$_{50:95}$ is then computed with the same standard COCO protocol used for all other models.}
While a perfect comparison is challenging due to different task designs, we make the input assumptions explicit in Figure~\ref{fig:vit_sam_inputs} InsCore-pretrained ViTDet takes only images as input, whereas SAM and SAM2 additionally receive bounding box prompts derived from ground truth.
InsCore-pretrained ViTDet therefore addresses the more challenging task of estimating both location and category from the entire image, and we report the prompted SAM/SAM2 results as reference points rather than as directly comparable baselines.

\subsection{Ablation study on InsCore}
\label{sec:ablation_inscore}
Before comparing our InsCore pre-training with conventional approaches, including VFMs, we first explore combinations of parameters for synthetic data generation and pre-training. In particular, as the main subject of this paper, we analyze representation learning under complex occlusion in Tables~\ref{tab:ablation_instances},~\ref{tab:ablation_occlusion_dense},~\ref{tab:ablation_type}.

\noindent\textbf{Number of instances per image (Table~\ref{tab:ablation_instances}).}
We adjust the number of instances per image as the parameter that most strongly influences occlusion in the synthetic dataset. Specifically, we vary the number of instances per image from 1 (no occlusion) up to 64, finding that performance saturates at around 32 to 64 instances.

\noindent\textbf{Relation between number of instances per image and instance center placement area (Table~\ref{tab:ablation_occlusion_dense}).}
We also examine a method for controlling occlusion density by adjusting the placement area for instance centers. As shown in Table~\ref{tab:ablation_occlusion_dense}, we jointly analyze the effect of the number of instances per image $\{16, 32, 64\}$ and placement area size to determine the optimal occlusion balance. The results show that placing 32 instances with centers sampled from a $1024 \times 1024$ region around the image center yields the best trade-off.

\noindent\textbf{Occlusion types (Table~\ref{tab:ablation_type}).}
We also evaluate different occlusion placement types. As described in Section~\ref{sec:occlusion_handling}, we vary instance center placement according to several distributions and spatial rules. Among these, fully random placement achieves the highest score; however, mixing all placement strategies yields comparable performance. We report both results in Table~\ref{tab:ablation_type}, and use the fully random strategy as the default setting for InsCore in the main experiments.

\noindent\textbf{Number of images for pre-training \hk{(Table~\ref{tab:ablation_datasize_images})}.} We also investigated the optimal number of synthetic images for InsCore pre-training by scaling the dataset size from 12.5k to 400k. We found that the performance improves as the data scales but saturates around 200k images, achieving the highest average mAP of 44.1 \hk{under the ablation schedule}, and slightly degrades at 400k; we discuss this saturation in Section~\ref{sec:toward_industrial_vfm}.
\hk{Note that Tables~\ref{tab:ablation_instances}--\ref{tab:ablation_datasize_images} use a 100-epoch pre-training schedule, whereas the InsCore entries in Table~\ref{tab:main_comparisons} use the final 200-epoch schedule with a global batch size of 8 (see Tables~\ref{tab:ablation_batchsize} and~\ref{tab:ablation_pretrain_epoch} in the Appendix), which accounts for the numerical difference between the two tables.}

\begin{table}[tb]
    \centering
    \begin{minipage}[t]{0.48\textwidth}
        \centering
        \caption{Ablation study on \#instances \hk{(\#ins)} per image.}
        \label{tab:ablation_instances}
        \resizebox{\linewidth}{!}{%
        \begin{tabular}{cccccc|c}
        \toprule[0.8pt]
        \hk{\#ins} & ES & LC & SN & IS & LS & Ave \\
        \midrule[0.5pt]
        1  & 22.1 & 11.5 & 61.9 & 21.3 & 93.2 & 42.0 \\
        2  & 22.1 & 11.7 & 62.2 & 22.3 & 93.8 & 42.4 \\
        4  & 22.9 & 12.3 & 62.0 & 21.9 & 93.9 & 42.6 \\
        8  & 23.5 & 11.7 & \textbf{62.3} & 22.3 & \textbf{94.5} & 42.8 \\
        16 & 22.5 & 12.7 & 61.9 & 21.2 & 93.8 & 42.4 \\
        32 & 22.1 & 13.4 & \textbf{62.3} & \textbf{22.4} & 93.8 & 42.8 \\
        64 & \textbf{24.1} & \textbf{13.6} & \textbf{62.3} & 21.5 & \textbf{94.4} & \textbf{43.1} \\
        \bottomrule[0.8pt]
        \end{tabular}%
        }
    \end{minipage}
    \hfill
    \begin{minipage}[t]{0.48\textwidth}
        \centering
        \caption{Ablation study on occlusion density.}
        \label{tab:ablation_occlusion_dense}
        \resizebox{\linewidth}{!}{%
        \begin{tabular}{cccccc|c}
        \toprule[0.8pt]
        \#ins / pos & ES & LC & SN & IS & LS & Ave \\
        \midrule[0.5pt]
        16 / 256  & 22.5 & 12.7 & 61.9 & 21.2 & 93.8 & 42.4 \\
        32 / 256  & 22.1 & 13.4 & \textbf{62.3} & 22.4 & 93.8 & 42.8 \\
        64 / 256  & 24.1 & 13.6 & \textbf{62.3} & 21.5 & \textbf{94.4} & 43.1 \\
        16 / 512  & 22.3 & 13.3 & 61.9 & 21.0 & 94.2 & 42.5 \\
        32 / 512  & 21.5 & 12.0 & 61.8 & 23.0 & 93.4 & 42.3 \\
        64 / 512  & 23.5 & 12.8 & 62.1 & 21.8 & 94.1 & 42.8 \\
        16 / 1024 & 22.6 & 13.5 & 62.0 & 21.9 & 93.9 & 42.7 \\
        32 / 1024 & \textbf{24.8} & \textbf{13.8} & \textbf{62.3} & \textbf{22.7} & 93.6 & \textbf{43.4} \\
        64 / 1024 & 24.0 & 12.3 & \textbf{62.3} & 22.1 & 94.1 & 42.9 \\
        \bottomrule[0.8pt]
        \end{tabular}%
        }
    \end{minipage}
\end{table}

\begin{table}[tb]
    \centering
    \begin{minipage}[t]{0.48\textwidth}
        \centering
        \caption{Ablation study on occlusion types.}
        \label{tab:ablation_type}
        \resizebox{\linewidth}{!}{%
        \begin{tabular}{cccccc|c}
        \toprule[0.8pt]
        Type & ES & LC & SN & IS & LS & Ave \\
        \midrule[0.5pt]
        Random          & \textbf{24.8} & \textbf{13.8} & 62.3 & 22.7 & 93.6 & \textbf{43.4} \\
        Grid            & 22.9 & 12.4 & 62.3 & 22.3 & 93.6 & 42.7 \\
        Gaussian        & 23.4 & 12.3 & 62.1 & 22.7 & 94.2 & 42.9 \\
        Poisson         & 23.0 & 12.0 & \textbf{62.4} & \textbf{23.6} & 93.7 & 42.9 \\
        Cluster         & 22.9 & 11.8 & \textbf{62.4} & 23.3 & 93.5 & 42.7 \\
        Spiral          & 23.5 & 12.4 & 62.1 & 21.9 & 94.0 & 42.7 \\
        Mixed all  & 23.1 & 12.3 & 62.3 & 23.2 & \textbf{94.4} & 43.1 \\
        \bottomrule[0.8pt]
        \end{tabular}%
        }
    \end{minipage}
    \hfill
    \begin{minipage}[t]{0.48\textwidth}
        \centering
        \caption{Ablation study on \#images at each pre-training dataset.}
        \label{tab:ablation_datasize_images}
        \resizebox{\linewidth}{!}{%
        \begin{tabular}{cccccc|c}
        \toprule[0.8pt]
        \#Images & ES & LC & SN & IS & LS & Ave \\
        \midrule[0.5pt]
        12.5k                  & \textbf{24.9} & \textbf{13.8} & 63.0 & 23.3 & 94.3 & 43.8 \\
        25k                    & 22.4 & 12.8 & 63.2 & \textbf{25.2} & \textbf{94.9} & 43.7 \\
        50k                    & 24.4 & 13.3 & 63.2 & 24.4 & 94.2 & 43.9 \\
        100k   & 24.8 & \textbf{13.8} & 62.3 & 22.7 & 93.6 & 43.4 \\
        200k                   & 24.4 & 12.7 & \textbf{63.4} & \textbf{25.2} & \textbf{94.9} & \textbf{44.1} \\
        400k                   & 24.2 & 12.6 & 63.2 & 24.0 & 94.2 & 43.6 \\
        \bottomrule[0.8pt]
        \end{tabular}%
        }
    \end{minipage}
\end{table}

\noindent\textbf{Optimized parameters \hk{(Table~\ref{tab:main_comparisons})}.}
Based on the parameter sweep described above, we use the following configuration for the main comparisons with the ViT-B backbone: 32 instances per image, a $1{,}024\times1{,}024$ pixel placement area, full random placement, 200k pre-training images, a global batch size of 8, and 200 pre-training epochs. \hk{We denote the model pre-trained with this optimized configuration (200k images) as InsCore* in Tables~\ref{tab:main_comparisons} and \ref{tab:comparison_vfm}.} The corresponding results are summarized in Table~\ref{tab:main_comparisons} \hk{; additional experiments are shown in the Appendix}.

\noindent\textbf{Swin Transformer (Table~\ref{tab:comparison_vfm}).}
We additionally experiment with a Swin-Base (Swin-B) backbone~\cite{liu2021swin}. For the Swin Transformer, using 100k pre-training images and 100 epochs yields better performance in our experiments.

\begin{table*}[t]
\centering
\caption{Comparison of the effectiveness of InsCore and other pre-training datasets on industrial segmentation tasks across five domains. The best and second-best scores for each dataset are indicated by \underline{\textbf{underlined bold}} and \textbf{bold} text, respectively. * denotes the InsCore pre-trained model with the optimized parameters and 200k pre-training images (Section~\ref{sec:ablation_inscore}).}
\label{tab:main_comparisons}
\begin{tabular}{lccccccccc}
\toprule[0.8pt]
PT data & Backbone & Image type & \#PT images & ES & LC & SN & IS & LS & Ave \\
\midrule[0.5pt]
Scratch & ViT-B & -- & -- & 16.6 & 11.9 & 59.5 & 12.1 & 92.2 & 38.5 \\
ImageNet-21k~\cite{deng2009imagenet} & ViT-B & Real & 14M & \underline{\textbf{29.1}} & \underline{\textbf{16.4}} & 61.1 & 24.1 & 94.2 & \textbf{45.0} \\
SegRCDB~\cite{shinoda2023segrcdb} & ViT-B & Synthetic & 0.1M & 23.0 & 13.0 & 61.3 & 21.9 & 94.1 & 42.6 \\
Hypersim~\cite{roberts2021hypersim} & ViT-B & Synthetic & 0.07M & 22.0 & \textbf{15.7} & 60.2 & 20.9 & 94.3 & 42.6 \\
Virtual KITTI~\cite{cabon2020virtual} & ViT-B & Synthetic & 0.02M & 19.1 & 14.7 & 59.9 & 17.3 & 94.3 & 41.1 \\
SAIL-VOS~\cite{hu2019sail} & ViT-B & Synthetic & 0.1M & 23.0 & 14.2 & 59.9 & 18.4 & \textbf{95.2} & 42.1 \\
\rowcolor[gray]{0.7}InsCore & ViT-B & Synthetic & 0.1M & 24.8 & 13.4 & \textbf{63.2} & \underline{\textbf{26.1}} & 94.9 & 44.5 \\
\rowcolor[gray]{0.7}InsCore* & ViT-B & Synthetic & 0.2M & \textbf{25.6} & 15.6 & \underline{\textbf{63.5}} & \underline{\textbf{26.1}} & \underline{\textbf{95.6}} & \underline{\textbf{45.2}} \\
\bottomrule[0.8pt]
\end{tabular}
\end{table*}

\begin{table}[t]
\centering
\caption{Pre-training effectiveness comparison between InsCore and segmentation foundation models on industrial datasets. SAM and SAM2 are prompted with ground-truth bounding boxes (with 1--3 pixel perturbations; see Table~\ref{tab:sam_head_bboxnoise_investigation}) and are class-agnostic, whereas the InsCore-pretrained models take only images as input; the SAM/SAM2 rows should therefore be read as reference points under different input assumptions. * denotes the optimized configuration (Section~\ref{sec:ablation_inscore}).}
\label{tab:comparison_vfm}
\resizebox{\textwidth}{!}{
\begin{tabular}{lccccccccccc}
\toprule[0.8pt]
Method & Backbone & Head & Prompt & \multicolumn{2}{c}{Pre-training} & \multicolumn{5}{c}{\hk{mAP}} & Ave \\
\cmidrule(lr){5-6} \cmidrule(lr){7-11}
 &  & & & Dataset & Size & ES & LC & SN & IS & LS &  \\ 
\midrule[0.5pt]
SAM~\cite{Kirillov_2023_ICCV} & ViT-B & -- & GT bbox & SA-1B & 11M & 48.7 & 9.3 & 52.3 & \textbf{36.4} & 80.3 & 45.4 \\
SAM2~\cite{ravi2024sam2} & ViT-B & -- & GT bbox & SA-1B+V & 11M+51k & \textbf{56.4} & 9.9  & 52.2 & 32.8 & 71.6 & 44.5 \\
\rowcolor[gray]{0.7}InsCore* & ViT-B & MRCNN & -- & InsCore & 0.2M & 25.6 & 15.6 & \textbf{63.5} & 26.1 & \textbf{95.6} & 45.2 \\
\rowcolor[gray]{0.7}InsCore & Swin-B & MRCNN & -- & InsCore & 0.1M & 29.7 & \textbf{18.3} & 61.6 & 25.5 & 95.1 & \textbf{46.0} \\
\bottomrule[0.8pt]
\end{tabular}
}
\end{table}

\subsection{Experimental results}
\label{sec:experimental_results}

\noindent\textbf{Comparison on pre-training datasets with real and synthetic images (Table~\ref{tab:main_comparisons}).}
For comparison, we include an ImageNet-21k pre-trained model and four synthetic segmentation datasets: Hypersim~\cite{roberts2021hypersim}, Virtual KITTI~\cite{cabon2020virtual}, SAIL-VOS~\cite{hu2019sail}, and SegRCDB~\cite{shinoda2023segrcdb}.
All models were fine-tuned on the five industrial datasets, and performance was measured using mean Average Precision (mAP$_{50:95}$), i.e., AP averaged over IoU thresholds from 0.50 to 0.95 in steps of 0.05.
As shown in Table~\ref{tab:main_comparisons}, InsCore* achieves the highest average mAP (45.2) across the five industrial datasets, on par with ImageNet-21k pre-training (45.0), while using no real images and clearly outperforming all existing synthetic pre-training datasets (41.1-42.6) under the same backbone and training recipe.
\hk{We position the comparison with ImageNet-21k as parity rather than superiority: per-domain results are mixed (ImageNet-21k is stronger on ES and LC, whereas InsCore is stronger on SN, IS, and LS), and when the nearly saturated LS (92.2 even from scratch) is excluded, both reach the same four-dataset average of 32.7. The key takeaway is that a fully synthetic, commercially usable dataset of only 0.2M images matches 14M supervised real images for industrial transfer.}

\noindent\textbf{Comparison with VFMs (Tables~\ref{tab:comparison_vfm} and \ref{tab:comparison_coco}).}
To further evaluate InsCore as a pre-training dataset, we report comparisons with SAM and SAM2.

As shown in Table~\ref{tab:comparison_vfm}, InsCore-pretrained ViTDet and Swin Transformer achieve 45.2 and 46.0 average mAP across the industrial datasets, respectively.
We note that SAM/SAM2 are prompt-based and class-agnostic, whereas ViTDet/Swin perform fully automatic instance segmentation from images alone;
therefore, these results should be interpreted under different input assumptions (see the Appendix and Figure~\ref{fig:vit_sam_inputs}).
These results indicate that InsCore, comprising only 0.1M-0.2M formula-generated synthetic images, can serve as an effective pre-training dataset in our setting, despite being substantially smaller than SA-1B (11M high-resolution real images).
\begin{wraptable}[10]{r}{0.48\textwidth}
    \centering
    \vspace{-20pt}
    \caption{Pre-training effectiveness comparison on MSCOCO~\cite{lin2014microsoft}.}
    \label{tab:comparison_coco}
    \setlength{\tabcolsep}{2pt}
    \small
    \resizebox{\linewidth}{!}{%
        \begin{tabular}{lcccc}
        \toprule[0.8pt]
        Method & Backbone & mAP & mAP$_{50}$ & mAP$_{75}$ \\
        \midrule[0.5pt]
        Scratch & ViT-B & 13.1 & 24.1 & 12.8 \\
        SAM~\cite{Kirillov_2023_ICCV} & ViT-B  & 40.4 & 60.4 & 44.0 \\
        \rowcolor[gray]{0.7}InsCore & ViT-B & 39.3 & 61.3 & 42.3 \\
        \rowcolor[gray]{0.7}InsCore & Swin-B & \textbf{44.4} & \textbf{68.2} & \textbf{47.5} \\
        \bottomrule[0.8pt]
        \end{tabular}%
    }
\end{wraptable}
Additionally, Table~\ref{tab:comparison_coco} reports results on MSCOCO for the InsCore-pretrained models and SAM.
With the ViT-B backbone, InsCore (39.3 mAP) remains slightly below SAM (40.4), which is consistent with InsCore's industrial-focused design, while the Swin-B backbone reaches 44.4.
These results suggest that dense-occlusion synthetic pre-training with pixel-accurate annotations transfers reasonably beyond the industrial datasets, although closing the remaining gap on general web-image benchmarks is not the goal of this work.

\section{Discussion: Toward a better industrial VFM}
\label{sec:toward_industrial_vfm}
We propose InsCore, a synthetic pre-training approach centered on learning complex occlusions for industrial instance segmentation.
Because InsCore uses only formula-generated synthetic images, it avoids licensing constraints associated with real-image data; in line with the FDSL family (e.g., FractalDB~\cite{kataoka2020pre}), both the generation code and the complete dataset will be released under a commercially usable license, ensuring extensibility and reproducibility for future industrial VFM research.
To conclude, we discuss key design considerations for better industrial VFMs.

\noindent\textbf{Computational efficiency and performance.}
Among the factors affecting pre-training efficiency, the number of pre-training images has a substantial impact (see Table~\ref{tab:ablation_datasize_images}).
The ViT-B backbone performs best with 200k images and 200 pre-training epochs (45.2 in Table~\ref{tab:main_comparisons}), whereas the Swin-B backbone performs best with 100k images and 100 epochs, achieving the highest average mAP in this paper (46.0).
In terms of scale, InsCore uses 100k images and 3.2M masks, approximately 110$\times$ and 312$\times$ fewer in images and mask labels than SA-1B (11M images, 1B masks).
\hk{We note that SA-1B was built to train general-purpose promptable segmentation models, whereas InsCore targets industrial pre-training; these numbers therefore indicate dataset scale rather than a like-for-like measure of data efficiency.}

\noindent\textbf{Performance at each industrial dataset.}
When prompted with perfect bounding boxes, SAM is strong on ES (medical) and IS (manufacturing), whereas InsCore-pretrained ViTDet outperforms it on LC (biomedical), SN (remote sensing), and LS (logistics); for SN and LS, the ViT-B backbone achieves the highest scores overall, indicating the potential of synthetic pre-training with ViT.

\noindent\textbf{Limitations and future work.}
InsCore's benefit is domain-dependent: on Endoscapes it falls behind ImageNet-21k (25.6 vs.\ 29.1), suggesting that hollow polygons rendered as white strokes may under-represent thin, elongated structures such as surgical instruments and provide limited texture cues for ambiguous tissue boundaries; enriching the shape and appearance diversity of the generator is a natural next step.
In addition, since occlusion handling is intertwined with dense, pixel-accurate annotation and boundary learning, isolating the individual contribution of each factor (including a controlled comparison of the hollow-mask design against solid or contour-only alternatives) remains future work, which the released generation code makes straightforward.
Finally, we report single-seed results, and performance saturates around 200k images (Table~\ref{tab:ablation_datasize_images}), indicating that further gains may require increasing generation diversity or combining InsCore with real images.

\begin{figure*}[!t]
    \centering
  \includegraphics[width=0.80\columnwidth]{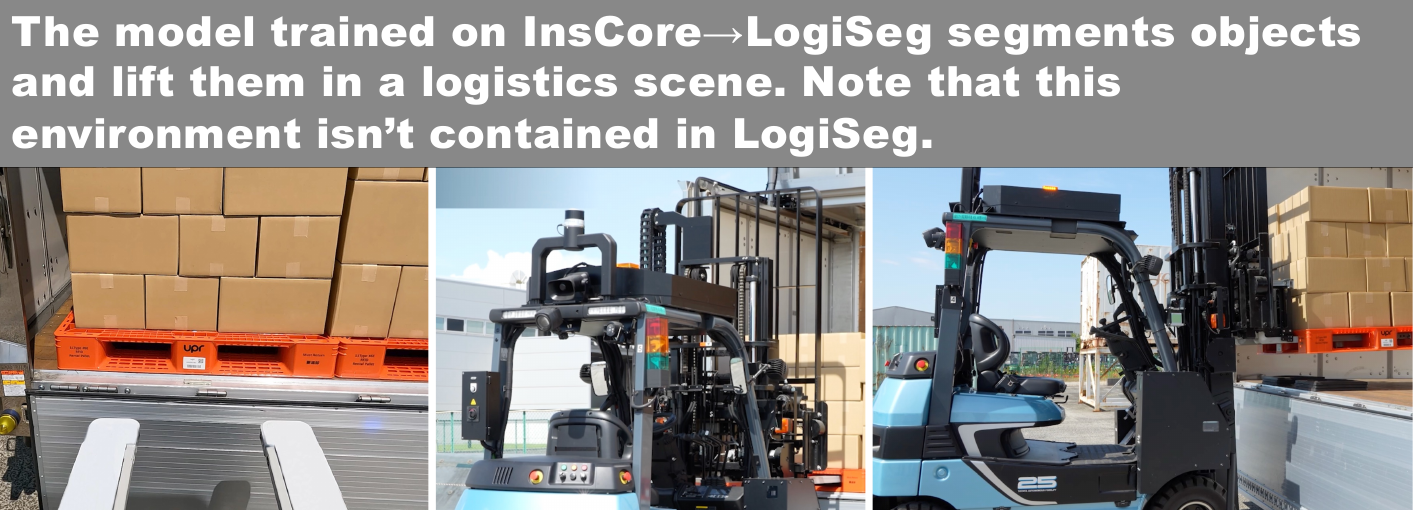}
  \caption{Autonomous forklift.}
  \label{fig:tech_transfer}
\end{figure*}

\noindent\textbf{Industrial application in logistics environments (Figure~\ref{fig:tech_transfer}).}
We deployed the InsCore pre-trained, LogiSeg (LS) fine-tuned model on an autonomous forklift. Predicted masks were projected into 3D via LiDAR-based 2D--3D calibration to estimate contact and clearance between the load and the pallet; in real-world validation, the system detected interference between loads as small as approximately 5\,mm, which we attribute to occlusion-aware pre-training with precisely annotated synthetic data.

\section*{Acknowledgement}

This work was supported by the AIST policy-based budget project ``R\&D on Generative AI Foundation Models for the Physical Domain''. This paper was supported by Japan Science and Technology Agency (JST) as part of Adopting Sustainable Partnerships for Innovative Research Ecosystem (ASPIRE), Grant Number JPMJAP2518. We used ABCI 3.0 provided by AIST and AIST Solutions with support from ``ABCI 3.0 Development Acceleration Use''.

\bibliographystyle{splncs04}
\bibliography{main}

\clearpage
\appendix
\renewcommand{\thetable}{A\arabic{table}}
\renewcommand{\thefigure}{A\arabic{figure}}
\setcounter{table}{0}
\setcounter{figure}{0}

\section{Appendix}
\label{sec:appendix}

\begin{figure*}[!t]
    \centering
    \includegraphics[width=0.95\textwidth]{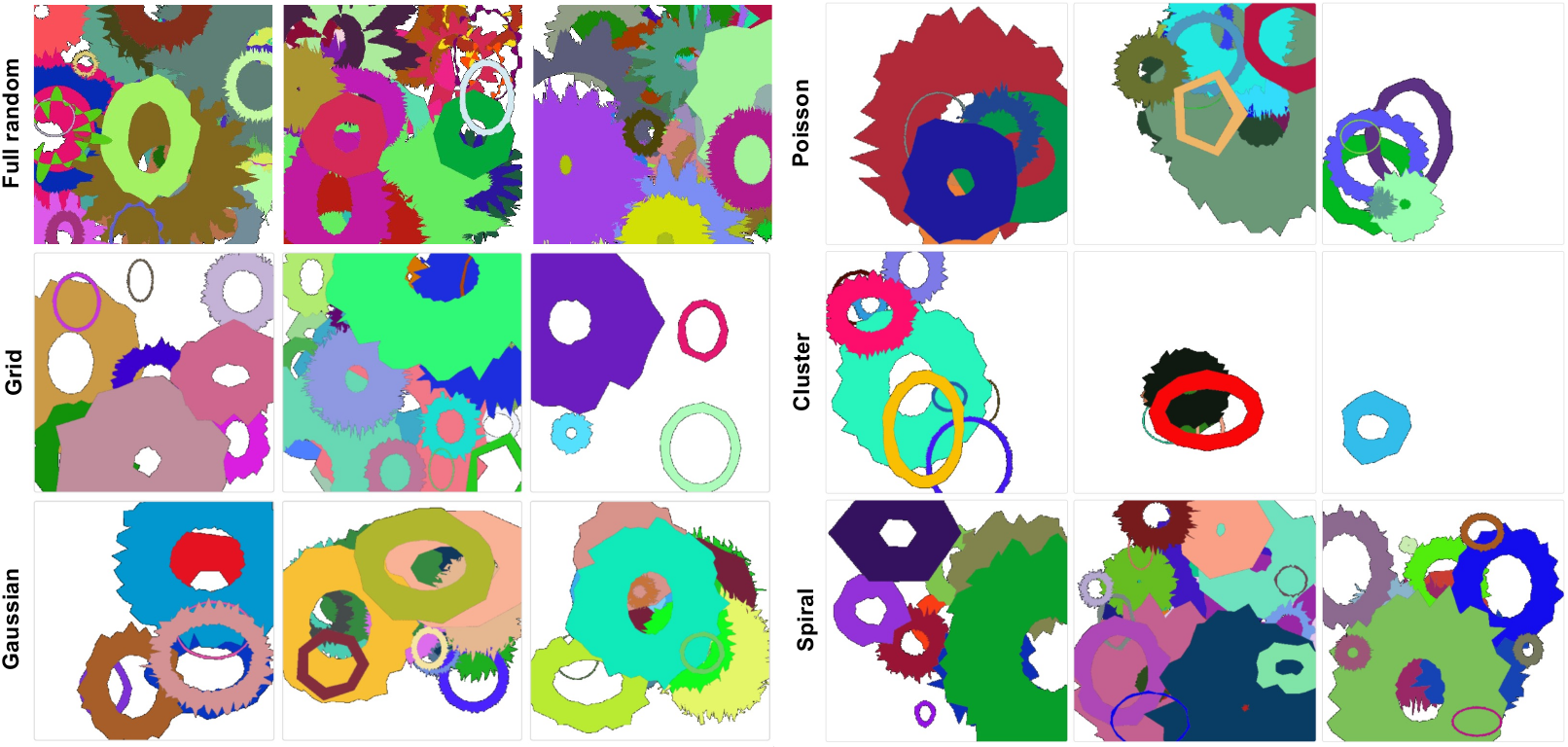}
    \caption{Visualization of the six occlusion types.}
    \label{fig:occlusion_types}
\end{figure*}

\noindent\textbf{Occlusion types (Figure~\ref{fig:occlusion_types}).}
See the figure for detailed visualizations of the six occlusion categories.

\noindent\textbf{Global batch size (Table~\ref{tab:ablation_batchsize}).}
We study the effect of the global batch size using mAP$_{50:95}$ on the five industrial datasets. Since our experiments use 8 GPUs, we set the lower bound of the batch size to 8. A batch size of 8 achieved the best average performance.

\noindent\textbf{Pre-training epochs (Table~\ref{tab:ablation_pretrain_epoch}).}
We also examine the effect of the number of pre-training epochs using mAP$_{50:95}$. Here, 0 epochs correspond to training from scratch. Compared with the 100-epoch setting, extending the pre-training to 200 epochs improves the average performance.

\noindent\textbf{Fine-tuning epochs (Table~\ref{tab:ablation_finetune}).}
We evaluated fine-tuning for 60, 100, and 200 epochs using mAP$_{50:95}$. The results show that training for 60 epochs achieved the highest average score, indicating that longer fine-tuning does not necessarily lead to better performance.

\noindent\textbf{How to construct SAM / SAM2 baselines (Table~\ref{tab:comparison_sam_finetuning} and Table~\ref{tab:sam_head_bboxnoise_investigation}).}
We first compare SAM with and without fine-tuning using ground-truth bounding-box prompts and mIoU in Table~\ref{tab:comparison_sam_finetuning}. Because fine-tuned SAM on LIVECell failed to optimize, Table~\ref{tab:comparison_sam_finetuning} reports results only on ES, SN, IS, and LS, together with the corresponding four-dataset average. The non-fine-tuned SAM yields the higher average score and is therefore used in subsequent comparisons.

Next, Table~\ref{tab:sam_head_bboxnoise_investigation} examines how we construct the prompted SAM / SAM2 baselines used in the main comparisons (Table~\ref{tab:comparison_vfm}). We also tested a SAM-based fine-tuning setting that uses SAM as the ViT-B backbone with a Mask R-CNN head and no prompts. However, this setting did not provide stable performance across datasets. We therefore use prompted SAM / SAM2 baselines. Specifically, we use ground-truth bounding boxes as prompts and add small random perturbations to each $(x, y)$ coordinate to model slight localization errors at inference time. Among the tested settings, box noise of 1--3 pixels gives the highest performance, and we adopt this setting as the SAM / SAM2 baseline.

\begin{table}[t]
\centering
\caption{Ablation study on global batch size, measured by mAP$_{50:95}$ on the five industrial datasets.}
\label{tab:ablation_batchsize}
\begin{tabular}{l|ccccc|c}
\toprule
\#Batch & ES & LC & SN & IS & LS & Average \\
\midrule
8  & \textbf{25.6} & 13.1 & \textbf{63.1} & \textbf{25.5} & 94.6 & \textbf{44.3} \\
16 & 24.3 & 12.5 & \textbf{63.1} & 23.4 & 94.5 & 43.5 \\
32 & 24.8 & 13.1 & \textbf{63.1} & 24.8 & \textbf{94.8} & 44.1 \\
64 & 23.2 & \textbf{13.2} & 62.6 & 23.4 & 94.2 & 43.3 \\
\bottomrule
\end{tabular}
\end{table}

\begin{table}[t]
\centering
\caption{Ablation study on pre-training epochs, measured by mAP$_{50:95}$ on the five industrial datasets.}
\label{tab:ablation_pretrain_epoch}
\begin{tabular}{l|ccccc|c}
\toprule
\#Epochs & ES & LC & SN & IS & LS & Average \\
\midrule
0   & 16.6 & 11.9 & 59.5 & 12.1 & 92.2 & 38.5 \\
50  & 24.1 & 13.6 & 62.3 & 21.5 & 94.4 & 43.2 \\
100 & 23.2 & 13.2 & 62.6 & 23.4 & 94.2 & 43.3 \\
200 & \textbf{24.8} & \textbf{13.4} & \textbf{63.2} & \textbf{26.1} & \textbf{94.9} & \textbf{44.5} \\
\bottomrule
\end{tabular}
\end{table}

\begin{table}[t]
\centering
\caption{Ablation study on fine-tuning epochs, measured by mAP$_{50:95}$ on the five industrial datasets.}
\label{tab:ablation_finetune}
\begin{tabular}{l|ccccc|c}
\toprule
\#Epochs & ES & LC & SN & IS & LS & Average \\
\midrule
60  & \textbf{24.8} & 13.4 & 63.2 & \textbf{26.1} & 94.9 & \textbf{44.5} \\
100 & 24.1 & \textbf{14.8} & \textbf{63.5} & 24.4 & 95.1 & 44.3 \\
200 & 24.0 & \textbf{14.8} & 63.2 & 24.6 & \textbf{95.2} & 44.3 \\
\bottomrule
\end{tabular}
\end{table}

\begin{table*}[t]
\centering
\caption{Comparison of SAM with and without fine-tuning using ground-truth bounding-box prompts. Results are reported in mIoU on ES, SN, IS, and LS. LIVECell is excluded from this table and from the average because fine-tuned SAM failed to optimize on that dataset. Mask R-CNN (Ours) is included as a reference. The best and second-best scores among the listed models for each dataset are indicated by \underline{\textbf{underlined bold}} and \textbf{bold} text, respectively.}
\label{tab:comparison_sam_finetuning}
\renewcommand{\arraystretch}{1.15}
\begin{tabular}{lccccccc}
\toprule
\multirow{2}{*}{Model} & \multirow{2}{*}{Backbone} & \multirow{2}{*}{Prompt} & \multicolumn{4}{c}{mIoU} & \multirow{2}{*}{Average} \\
\cmidrule(lr){4-7}
& & & ES & SN & IS & LS & \\
\midrule
SAM (w/o fine-tuning) & \multirow{2}{*}{ViT-B} & \multirow{2}{*}{GT bbox} & \textbf{65.8} & \textbf{75.1} & 52.0 & 90.0 & \textbf{70.7} \\
SAM (w/ fine-tuning)  & & & 51.3 & 73.0 & \textbf{60.6} & \textbf{90.2} & 68.8 \\
\midrule
\rowcolor[gray]{0.9} Mask R-CNN (Ours) & Swin-B & -- & \underline{\textbf{66.0}} & \underline{\textbf{76.9}} & \underline{\textbf{60.8}} & \underline{\textbf{96.4}} & \underline{\textbf{75.0}} \\
\bottomrule
\end{tabular}
\end{table*}

\begin{table*}[t]
\centering
\caption{SAM/SAM2 investigations for baseline construction. Comparison between SAM and SAM2 under different settings and noise levels. The prompted SAM/SAM2 variants use ground-truth bounding boxes with small random perturbations.}
\label{tab:sam_head_bboxnoise_investigation}
\resizebox{\textwidth}{!}{%
\begin{tabular}{lccccccccc|c}
\toprule
Model & PT & Backbone & Setting & ES & LC & SN & IS & LS & Average & Desc. \\
\midrule
SAM  & SA-1B   & ViT-B & MRCNN head    & 3.1  & 0.6  & \textbf{59.7} & 30.5 & 14.3 & 21.6 & No prompt \\
\midrule
SAM  & SA-1B   & ViT-B & GT bbox prompt & 48.7 & 9.3  & 52.3 & \textbf{36.4} & \textbf{80.3} & \textbf{45.4} & box noise (1--3 px) \\
SAM  & SA-1B   & ViT-B & GT bbox prompt & 48.2 & 7.6  & 50.0 & 30.5 & 79.8 & 43.2 & box noise (1--5 px) \\
SAM  & SA-1B   & ViT-B & GT bbox prompt & 44.1 & 2.2  & 34.4 & 10.3 & 75.7 & 33.4 & box noise (5--10 px) \\
SAM  & SA-1B   & ViT-B & GT bbox prompt & 27.5 & 0.3  & 3.4  & 3.6  & 56.7 & 18.3 & box noise (10--30 px) \\
\midrule
SAM2 & SA-1B+V & ViT-B & GT bbox prompt & \textbf{56.4} & \textbf{9.9}  & 52.2 & 32.8 & 71.6 & 44.5 & box noise (1--3 px) \\
SAM2 & SA-1B+V & ViT-B & GT bbox prompt & 56.3 & 8.0  & 50.0 & 27.0 & 71.3 & 42.5 & box noise (1--5 px) \\
\bottomrule
\end{tabular}%
}
\end{table*}

\clearpage

\begin{figure*}[!t]
    \centering
    \includegraphics[width=0.95\textwidth]{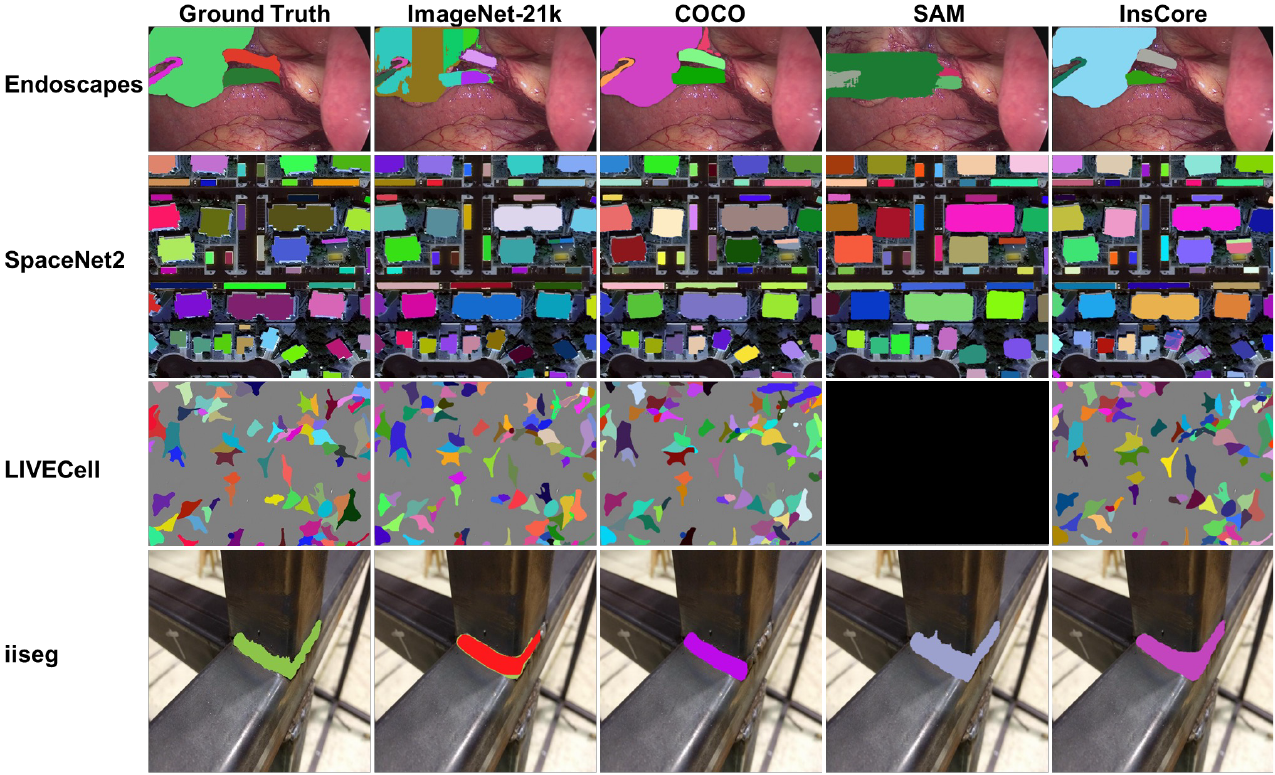}
    \caption{Qualitative instance segmentation results on Endoscapes (ES), SpaceNet2 (SN), LIVECell (LC), and Industrial-iSeg (IS). Each row corresponds to a dataset, and each column shows ground truth, ImageNet-21k, COCO, the SAM-based fine-tuned model, and InsCore. COCO is included as an additional qualitative reference. The SAM-based model failed to optimize on LIVECell; therefore, the corresponding panel is blacked out.}
    \label{fig:result_sample}
\end{figure*}

\noindent\textbf{Qualitative evaluation on instance segmentation.}
We qualitatively evaluate the segmentation performance of the InsCore-pre-trained ViTDet on industrial datasets. Figure~\ref{fig:result_sample} presents instance segmentation results after fine-tuning on four datasets: Endoscapes (ES), LIVECell (LC), SpaceNet2 (SN), and Industrial-iSeg (IS). We compare ImageNet-21k and InsCore, and also include a COCO pre-trained model as an additional qualitative reference. The SAM column corresponds to the SAM-based fine-tuned model that uses SAM as the ViT-B backbone with a Mask R-CNN head, rather than the prompted SAM baseline used in the quantitative comparison (Table~\ref{tab:comparison_vfm}). This SAM-based model failed to optimize on LC; therefore, the corresponding panel is blacked out in the figure. Ground-truth annotations are also included for reference.

InsCore consistently produces masks closely aligned with the ground truth. In ES, it better separates instruments from tissue. In SN, it reconstructs building contours with high accuracy. On the challenging LC dataset, with dense overlapping cells, InsCore achieves superior separation. In IS, it yields smoother boundaries and fewer false positives than ImageNet-21k and COCO. LogiSeg (LS), also evaluated in Table~\ref{tab:comparison_vfm}, is excluded from the figure due to legal restrictions. These qualitative results are consistent with the quantitative comparisons in Section~\ref{sec:experimental_results} and show that InsCore produces accurate instance masks across diverse industrial domains.

\end{document}